# Creating Prototypes for Fast Classification in Dempster-Shafer Clustering

Johan Schubert

Department of Information System Technology,
Division of Command and Control Warfare Technology,
Defence Research Establishment,
SE-172 90  Stockholm, SWEDEN
E-mail: schubert@sto.foa.se

**Abstract.** We develop a classification method for incoming pieces of evidence in Dempster-Shafer theory. This methodology is based on previous work with clustering and specification of originally nonspecific evidence. This methodology is here put in order for fast classification of future incoming pieces of evidence by comparing them with prototypes representing the clusters, instead of making a full clustering of all evidence. This method has a computational complexity of $O(M \cdot N)$ for each new piece of evidence, where $M$ is the maximum number of subsets and $N$ is the number of prototypes chosen for each subset. That is, a computational complexity independent of the total number of previously arrived pieces of evidence. The parameters $M$ and $N$ are typically fixed and domain dependent in any application.

## 1   Introduction

In this paper we develop a classification method for incoming pieces of evidence in Dempster-Shafer theory [6]. This methodology is based on earlier work [1] where we investigate a situation where we are reasoning with multiple events that should be handled independently. The approach was, that when we received several pieces of evidence about different and separate events we classified them according to which event they were referring to. That is, we made a partitioning of all sources of evidence into subsets, where all sources of evidence in a given subset concern one of the multiple events, and there is a one-to-one correspondence between the events and the elements of the partitioning. This was done in such a way that the overall conflict across all the subsets was minimal. By a source of evidence we typically mean some kind of sensor generating pieces of evidence, and not an object discovered by the sensor.

The aim of the partitioning is to cluster the sources of evidence in subsets, and select a limited number of prototypical sources of evidence for future classification, in order to speed up computation.

In figure 1 these subsets are denoted by $\chi_i$ and the conflict in $\chi_i$ is denoted by $c_i$. Here, thirteen pieces of evidence are partitioned into four subsets. When the number of subsets is uncertain there will also be a "domain conflict" $c_0$ which is a conflict between the current number of subsets and domain knowledge. The partition is then simply an allocation of all evidence to the different events.

If it is uncertain to which event some evidence is referring we have a problem. It could then be impossible to know directly if two different pieces of evidence are referring to the same event. We do not know if we should put them into the same subset or not.

To solve this problem, we can use the conflict in Dempster's rule when all evidence within a subset are combined, as an indication of whether these pieces of evidence belong together. The higher this conflict is, the less credible that they belong together.

Let us create an additional piece of evidence for each subset where the proposition of this additional evidence states that this is not an "adequate partition". Let the proposition take a value equal to the conflict of the combination within the subset. These new pieces of evidence, one regarding each subset, reason about the partitioning of the original evidence. Just so we do not confuse them with the original evidence, let us call these pieces of evidence "metalevel evidence" and let us say that their combination and the analysis of that combination take place on the "metalevel", figure 1.

In the combination of all metalevel evidence we only receive support stating that this is not an "adequate partition". We may call this support a "metaconflict". The smaller this support is, the more credible the partitioning. Thus, the most credible partitioning is the one that minimizes the metaconflict.

In [2] we further investigated the consequence of transferring different pieces of evidence between the subsets. This was done by observing changes in conflict when we moved a piece of evidence from one subset to another. Such changes in conflict were also interpreted as "metalevel evidence" indicating that the piece of evidence was misplaced.

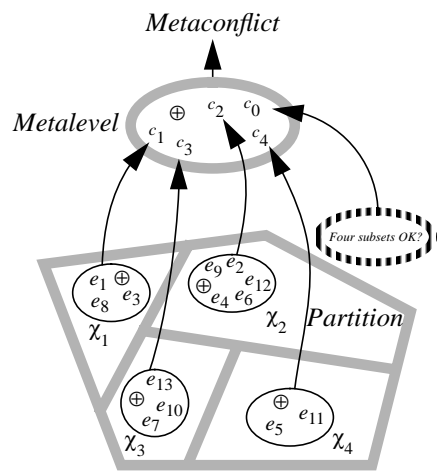

**Fig. 1.** The conflict in each subset of the partitioning becomes a piece of evidence at the metalevel.

Based on these items of metalevel evidence we are in this paper able to find potential prototypes as representatives for the subsets.

With the metalevel evidence we can make a partial specification of each original piece of evidence.

With such a partial specification for the potential prototypes we can find a measure of credibility for the correct classification of these prototypes. That credibility is the basis for choosing a fixed number of prototypes among all the potential prototypes as representatives for the subsets.

With the prototypes we will be able to do fast classification of all incoming new items of evidence. The reason why this approach is faster than the previous method in [1] for separating evidence that is mixed up, is that we now make only a simple comparison between an incoming item of evidence and the prototypes of the different subsets. In [1] we did a complete reclustering each time we received some new evidence in order to find a new partitioning, and thus the classification of the newly arrived piece of evidence.

The approach described corresponds in some sense to an idea by T. Denœux [7]. In that article Denœux assumes that a collection of preclassified prototypes are available and compares incoming evidence with the prototypes. He then defines a new item of evidence for each incoming piece of evidence and every prototype. This new item of evidence has a proposition which states that the incoming piece of evidence belongs to the same subset as the prototype. Combining all this new evidence yields the classification.

However, in this paper, we only use prototypes in order to obtain faster classification. These prototypes are not assumed to be preclassified outside of this methodology, instead their classification is derived from a previous clustering process [1]. It is quite possible to cluster and thus classify all incoming evidence without any prototypes as has been done in earlier work [1-5]. While the advantage with this approach is obvious, a fast computation time, there is an disadvantage in that future classification will have a correctness that might by less that what would have been possible if all previous evidence was used in the classification process and not only those that was obtained prior to clustering process.

Of course, it will always be possible to combine the two approaches of this paper and of earlier articles in such a way that the approach in this paper is used in a front process where time-critical calculation is performed and the usual clustering process is made in a back process where we have no time considerations. This way we may always obtain fast classification without suffering any long term degradation in performance.

In Section 2 we review two previous articles [1-2]. They form the foundation for the creation of prototypes in Section 3. A more extensive summary of [1-2] is found in [4]. In Section 4 we develop a method for fast classification based on these prototypes. This method is $O(M \cdot N)$, where $M$ is the maximum number of subsets allowed by an apriori probability distribution regarding the number of subsets, and $N$ is the number of prototypes chosen for each subset. Finally, in Section 5 we draw conclusions.

## 2  A Summary of Articles [1-2]

In an earlier article [1] we derived a method, within the framework of Dempster-Shafer theory [6], to handle evidence that is weakly specified in the sense that it may not be certain to which of several possible events a proposition is referring. If we receive such evidence about different and separate events and the pieces of evidence are mixed up, we want to classify them according to event.

In this situation it is impossible to directly separate pieces of evidence based only on their proposition. Instead we can use the conflict in Dempster's rule when all pieces of evidence within a subset are combined, as an indication of whether they belong together. The higher this conflict is, the less credible it is that they belong together.

### 2.1  Separating Nonspecific Evidence

In [1] we established a criterion function of overall conflict called the metaconflict function. With this criterion we can partition the evidence into subsets, each subset representing a separate event. We will use the minimizing of the metaconflict function as the method of partitioning. This method will also handle the situation when the number of events are uncertain.

#### 2.1.1  Metaconflict as a Criterion Function

Let $E_i$ be a domain proposition stating that there are $i$ subsets, $\Theta_E = \{E_0, ..., E_n\}$ a frame of such propositions and $m(E_i)$ the support for proposition $E_i$.

The metaconflict function can then be defined as:

DEFINITION. *Let the* metaconflict function,

$$Mcf(r, e_1, e_2, ..., e_n) \triangleq 1 - (1 - c_0) \cdot \prod_{i=1}^{r} (1 - c_i),$$

*be the conflict against a partitioning of n pieces of evidence into r disjoint subsets $\chi_i$ where $c_i$ is the conflict in subset i and $c_0$ is the conflict between the hypothesis that there are r subsets and our prior belief about the number of subsets.*

For a fixed number of subsets a minimum of the metaconflict function can be found by an iterative optimization among partitionings. In each step of the optimization the consequence of transferring a piece of evidence from one subset to another is investigated.

## 2.2 Specifying Nonspecific Evidence

In [2] we went further by specifying each piece of evidence by observing changes in cluster and domain conflicts if we move a piece of evidence from one subset to another, figure 2.

### 2.2.1 Evidence About Evidence

A conflict in a subset $\chi_i$ is interpreted as a piece of metalevel evidence that there is at least one piece of evidence that does not belong to the subset;

$$m_{\chi_i}(\exists j . e_j \notin \chi_i) = c_i.$$

Note that the proposition always takes a negative form $e_j \notin \chi_i$.

If a piece of evidence $e_q$ in $\chi_i$ is taken out from the subset, the conflict $c_i$ in $\chi_i$ decreases to $c_i^*$. This decrease $c_i - c_i^*$ is interpreted as metalevel evidence indicating that $e_q$ does not belong to $\chi_i$, $m_{\Delta\chi_i}(e_q \notin \chi_i)$, and the remaining conflict $c_i^*$ is an other piece of metalevel evidence indicating that there is at least one other piece of evidence $e_j$, $j \neq q$, that does not belong to $\chi_i - \{e_q\}$,

$$m_{\chi_i - \{e_q\}}(\exists j \neq q . e_j \notin (\chi_i - \{e_q\})) = c_i^*.$$

The unknown bpa, $m_{\Delta\chi_i}(e_q \notin \chi_i)$, is derived by stating that the belief that there is at least one piece of evidence that does not belong to $\chi_i$ should be the same irrespective of whether that belief is based on the original metalevel evidence before $e_q$ is taken out from $\chi_i$, or is based on a combination of the other two pieces of metalevel evidence after $e_q$ is taken out from $\chi_i$.

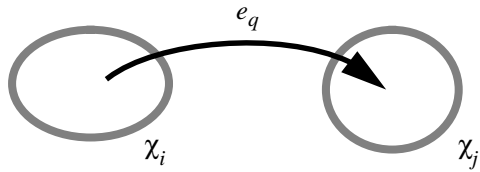

**Fig. 2.** Moving a piece of evidence changes the conflicts in $\chi_i$ and $\chi_j$.

We derive the metalevel evidence that $e_q$ does not belong to $\chi_i$ as

$$m_{\Delta\chi_i}(e_q \notin \chi_i) = \frac{c_i - c_i^*}{1 - c_i^*}.$$

We may calculate $m_{\Delta\chi_i}(e_q \notin \chi_i)$ for every $\chi_i$, figure 3.

If $e_q$ after it is taken out from $\chi_i$ is brought into another subset $\chi_k$, its conflict will increase from $c_k$ to $c_k^*$. The increase in conflict is interpreted as if there exists some metalevel evidence indicating that $e_q$ does not belong to $\chi_k + \{e_q\}$, i.e.,

$$\forall k \neq i . m_{\Delta\chi_k}(e_q \notin (\chi_k + \{e_q\})) = \frac{c_k^* - c_k}{1 - c_k}.$$

When we take out a piece of evidence $e_q$ from subset $\chi_i$ and move it to some other subset we might see a change in domain conflict. This indicates that the number of subsets is incorrect.

Here, we receive

$$m_{\Delta\chi}(e_q \notin \chi_{n+1}) = \frac{c_0^* - c_0}{1 - c_0}, |\chi_i| > 1,$$

$$m_{\Delta\chi}(e_q \notin \chi_i) = \frac{c_0 - c_0^*}{1 - c_0^*}, |\chi_i| = 1, c_0^* < c_0, \text{ and } m_{\Delta\chi}(e_q \in \chi_i) = \frac{c_0}{c_0^*}, |\chi_i| = 1, c_0^* > c_0.$$

### 2.2.2 Specifying Evidence

We may now make a partial specification of each piece of evidence. We combine all metalevel evidence from different subsets regarding a particular piece of evidence and calculate for each subset the belief and plausibility that this piece of evidence belongs to the subset.

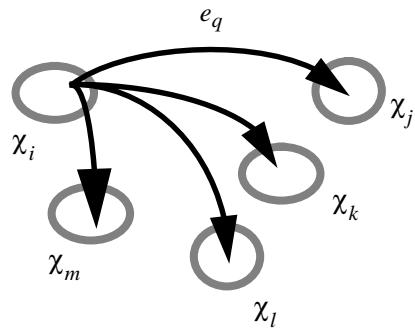

**Fig. 3.** Test moving $e_q$ to every other subset.

531

For the case when $e_q$ is in $\chi_i$ and $|\chi_i| > 1$ we receive, for example,

$$\forall k \neq n+1. \text{Bel}(e_q \in \chi_k) = 0$$

and

$$\forall k \neq n+1. \text{Pls}(e_q \in \chi_k) = \frac{1 - m(e_q \notin \chi_k)}{1 - \prod_{j=1}^{n+1} m(e_q \notin \chi_j)}.$$

### 2.3 Finding Usable Evidence

Some pieces of evidence might belong to one of several different subsets. Such an item of evidence is not so useful and should not be allowed to strongly influence a subsequent reasoning process.

If we plan to use a piece of evidence in the reasoning process of some subset, we must find a credibility that it belongs to the subset in question. A piece of evidence that cannot possible belong to a subset has a credibility of zero for that subset, while a piece of evidence which cannot possibly belong to any other subset and is without any support whatsoever against this subset has a credibility of one. That is, the degree to which some piece of evidence can belong to a certain subset and no other, corresponds to the importance it should be allowed to carry in that subset.

The credibility $\alpha_j$ of $e_q$ when $e_q$ is used in $\chi_j$ can then be calculated as

$$\alpha_j = [1 - \text{Bel}(e_q \in \chi_i)] \cdot \frac{[\text{Pls}(e_q \in \chi_j)]^2}{\sum_k \text{Pls}(e_q \in \chi_k)}, j \neq i,$$

$$\alpha_i = \text{Bel}(e_q \in \chi_i) + [1 - \text{Bel}(e_q \in \chi_i)] \cdot \frac{[\text{Pls}(e_q \in \chi_i)]^2}{\sum_k \text{Pls}(e_q \in \chi_k)}.$$

Here, $\text{Bel}(e_q \in \chi_i)$ is equal to zero except when $e_q \in \chi_i$, $|\chi_i| = 1$ and $c_0 < c_0^*$.

In [3] this work was further extended to find also a posterior probability distribution regarding the number of clusters.

## 3 Creating Prototypes

Each piece of evidence is a potential prototype for its most credible subset. When searching for the highest credibility $\alpha_j$ we notice that the term

$$\sum_k \text{Pls}(e_q \in \chi_k)$$

is independent of $j$. Let us call this term $K_q$.

When $\text{Bel}(e_q \in \chi_i)$ is equal to zero we may rewrite $\alpha_j$ as

$$\alpha_j = \frac{1}{K_q} \cdot [\text{Pls}(e_q \in \chi_j)]^2 = \frac{1}{K_q} \cdot [1 - m(e_q \notin \chi_j)]^2$$

We see that finding the maximum credibility $\alpha_j$ for $e_q$ is equal to finding the minimum evidence against the proposition that $e_q \in \chi_j$, i.e., maximizing $\alpha_j$ for $e_q$ and all $j$ is equal to minimizing $m(e_q \notin \chi_j)$ for $e_q$ and all $j$.

If $\text{Bel}(e_q \in \chi_i) \neq 0$ we have found earlier that $\text{Pls}(e_q \in \chi_i) = 1$, [2] Section III, and thus $\alpha_i > \alpha_j$ for $j \neq i$.

We have

$$\alpha_i = \text{Bel}(e_q \in \chi_i) + [1 - \text{Bel}(e_q \in \chi_i)] \cdot \frac{[\text{Pls}(e_q \in \chi_i)]^2}{\sum_k \text{Pls}(e_q \in \chi_k)}$$

$$= \text{Bel}(e_q \in \chi_i) + [1 - \text{Bel}(e_q \in \chi_i)] \cdot \frac{1}{K_q}$$

with $K_q \geq 1$ because $\text{Pls}(e_q \in \chi_i) = 1$.

Since

$$\text{Bel}(e_q \in \chi_i) = m(e_q \in \chi_i) + [1 - m(e_q \in \chi_i)] \cdot \prod_{\chi_j \in \chi^{-i}} m(e_q \notin \chi_j),$$

(see [2] Appendix II.A.3), and

$$\prod_{\chi_j \in \chi^{-i}} m(e_q \notin \chi_j) \leq 1$$

where $\chi^{-i} = \chi - \{\chi_i\}$ and $\chi = \{\chi_1, ..., \chi_n\}$, we find that the maximum value of $\alpha_i$ is obtained when $m(e_q \in \chi_i)$ is maximal. We have $\alpha_i = 1$ when $m(e_q \in \chi_i) = 1$.

A simple decision rule is then:

For every piece of evidence $e_q$

1. For $e_q$ and all $j$ find the maximum $m(e_q \in \chi_j)$. If the maximum $m(e_q \in \chi_j) \neq 0$ then $e_q$ is a potential prototype for $\chi_j$.

    2. However, if the maximum $m(e_q \in \chi_j) = 0$ then for $e_q$ and all $j$ find the minimum $m(e_q \notin \chi_j)$. Now, we have $e_q$ as a potential prototype for $\chi_j$.

However, we must use the credibility itself when determining which pieces of evidence among the potential prototypes for a certain subset will actually by chosen as one of the $N$ prototypes for that subset. While the above approach chooses the best subset $\chi_j$ for each piece of evidence by maximizing $m(e_q \in \chi_j)$ or minimizing $m(e_q \notin \chi_j)$, it is still possible that the evidence might be quite useless as a prototype since it could almost have been a potential prototype for some other subset. By ranking the potential prototypes within each subset according to credibility, we are able to find the most appropriate ones.

Note that in [2] we stated that in a subsequent reasoning process within a subset we could use each piece of evidence that had a credibility above zero for that subset, provided that we discounted it to its credibility.

We could then use almost every piece of evidence within each subset and we would get small contributions to the characteristics of the subset from pieces of evidence that most likely does not belong to the subset. A discounted piece of evidence might have something to offer in the reasoning process of a subset even when it is discounted to its credibility, but we do not know for sure that it belongs to the subset.

Thus, a substantially discounted piece of evidence hardly makes an ideal candidate as a prototype for a subset. From the prototypes we want to find the characteristics of the subset. That is something that cannot be guaranteed if we do not know with a high credibility that the piece of evidence belongs to the subset.

Also, the computational complexity would become exponential when making future classification of incoming pieces of evidence, since the number of focal elements in the result of the final combination of all pieces of evidence within the subset, would grow exponentially with the number of pieces of evidence within the subset. That means that we would get an exponential-time algorithm when classifying future incoming pieces of evidence even if all present pieces of evidence within the subset where combined in advance.

Instead we choose the $N$ prototypes with the highest credibility for the subset. They are judged to be the best representatives of the characteristics of the subset. Our future classification will be based on a comparison between them and the new incoming piece of evidence.

A second decision rule can then be formulated as:
For every subset
    1. Of the potential prototypes allocated for a subset, choose the $N$ prototypes with highest credibility for that subsets as the actual prototypes.
    2. Disregard the other potential prototypes for that subset.

By first applying the first decision rule for all pieces of evidence and then the second decision rule for every subset we are able to find $N$ different prototypes for each subset provided, of course, that there are at least $N$ potential prototypes for each subset.

Finally, we combine all prototypes within each subset into one new basic probability assignment. This way, each subset will now contain only one piece of evidence. While doing that, we also make a note of the conflict $c_j$ received in that combination. We will need it in the fast classification process.

## 4   Fast Classification

Now, given that we have all the prototypes for each subset, we can make fast classification of future incoming pieces of evidence. We will use the derived items of metalevel evidence $m_{\Delta\chi_i}(e_q \notin \chi_i)$.

If the evidence for $e_q$ against every subset is very high we will not classify $e_q$ as belonging to any of the subsets $\chi_j$, $j \leq n$. We will use a rejection rule if the best subset for $e_q$ is no better than it would be to create an additional subset $\chi_{n+1}$, and take the penalty for that from an increase in domain conflict.

Our rejection rule is:

Reject $e_q$ if the minimum for all $j$ of $m_{\Delta\chi_j}(e_q \notin \chi_j)$ is larger than $m_{\Delta\chi}(e_q \notin \chi_{n+1})$ where

$$m_{\Delta\chi_j}(e_q \notin \chi_j) = \frac{c_j^* - c_j}{1 - c_j} \text{ and } m_{\Delta\chi}(e_q \notin \chi_{n+1}) = \frac{c_0^* - c_0}{1 - c_0}.$$

With

$$c_0 = \sum_{i \neq n} m(E_i) \text{ and } c_0^* = \sum_{i \neq n+1} m(E_i),$$

$m(E_i)$ being the prior support given to the fact there are $i$ subsets, we have

$$m_{\Delta\chi}(e_q \notin \chi_{n+1}) = \frac{c_0^* - c_0}{1 - c_0} = \frac{\sum_{i \neq n+1} m(E_i) - \sum_{i \neq n} m(E_i)}{1 - \sum_{i \neq n} m(E_i)} = \frac{m(E_{n+1}) - m(E_n)}{1 - \sum_{i \neq n} m(E_i)},$$

where $m(E_{n+1})$ is always greater than $m(E_n)$, [1].

If $e_q$ is not rejected by this rule, then $e_q$ is classified as belonging to the subset $\chi_j$ for which $m_{\Delta\chi_j}(e_q \notin \chi_j)$ is minimal for all $j$.

All it takes to find $c_j^*$ is one combination of Dempster's rule for each cluster between the incoming piece of evidence and the already made combination of the prototypes of that cluster. We already have $c_j$ for every subset as well as $m(E_i)$ for all $i$.

If a fixed maximum number of prototypes *N* are used for each cluster and an apriori probability distribution regarding the number of clusters has some maximum number of subsets *M*, i.e., m($E_i$) = 0 if $i > M$, then the classification can always be done in time $O(M \cdot N)$. That is, independent of the total number of pieces of evidence in the previous clustering process.

## 5  Conclusions

In this paper we have shown how to create a fixed number of prototypes for each subset. They, and only they, will be the representatives of that subset. If there is also a limit on the maximum number of clusters through the apriori distribution regarding the number of subsets, then we can do a fast classification of all future incoming pieces of evidence. The performance of this methodology will obviously vary with the maximum number of prototypes *N*. If *N* is too large the computation time will become too long although it will grow only linearly in *N*. If on the other hand *N* it is too small, then the classification process will make too many classification errors. The actual choice of the parameter *N* will be domain dependent and has to be found for each application separately.